\documentclass[letterpaper, 10 pt, conference]{ieeeconf}  

\usepackage{graphicx}
\usepackage{multirow}
\usepackage{amsmath}
\usepackage{amssymb}
\usepackage{array}
\usepackage{subcaption}
\usepackage[dvipsnames]{xcolor}
\usepackage{color,soul}
\usepackage{url}
\usepackage{booktabs}
\usepackage{cite}
\usepackage[font=small]{caption}

\IEEEoverridecommandlockouts                              

\title{\LARGE \bf
Reinforcement Learning Based Heterogeneous Task Allocation Under Spatio-Temporal Environmental Uncertainty for Multi-Robot Teams 
}

\author{Xi Lin, Yaxuan Li, Paul Szenher and Brendan Englot
\thanks{Xi Lin, Yaxuan Li, Paul Szenher and Brendan Englot are with the Department of Mechanical Engineering, Stevens Institute of Technology, 1 Castle Point Terrace, Hoboken, NJ 07030, USA,
{\tt\small \{xlin26,yli21,pszenher,benglot\}@stevens.edu}
\;\;\; This work was supported by ONR Grant N00014-24-1-2522.}
}

\begin{document}

\maketitle
\thispagestyle{empty}
\pagestyle{empty}

\begin{abstract}
Autonomous Surface Vehicles (ASVs) have been increasingly deployed to perform different kinds of maritime tasks in recent years, which motivates the need for a reliable task allocation policy that can adapt to heterogeneous tasks and maritime disturbances.
In this work, we formulate a multi-robot task allocation problem in which a team of ASVs with heterogeneous capabilities needs to complete tasks of different types and appearance times under unknown stochastic spatio-temporal uncertainties from maritime disturbances.
We propose a decentralized Reinforcement Learning (RL) based policy network that is trained to be robust towards uncertainty-induced observation deviations, and is able to perform individual decision making that leads to efficient execution behaviors toward heterogeneous task requirements.
We evaluate the proposed policy and four baseline methods in simulated experiments with 
varied
proportions of different tasks types, numbers of tasks, robot species and robots per species.
Experiment results show that our proposed policy achieves the lowest makespan and can be deployed to address scenarios with 100 robots and 400 tasks. 
\end{abstract}

\section{Introduction}
\begin{figure}
    \centering
    \includegraphics[width=\linewidth]{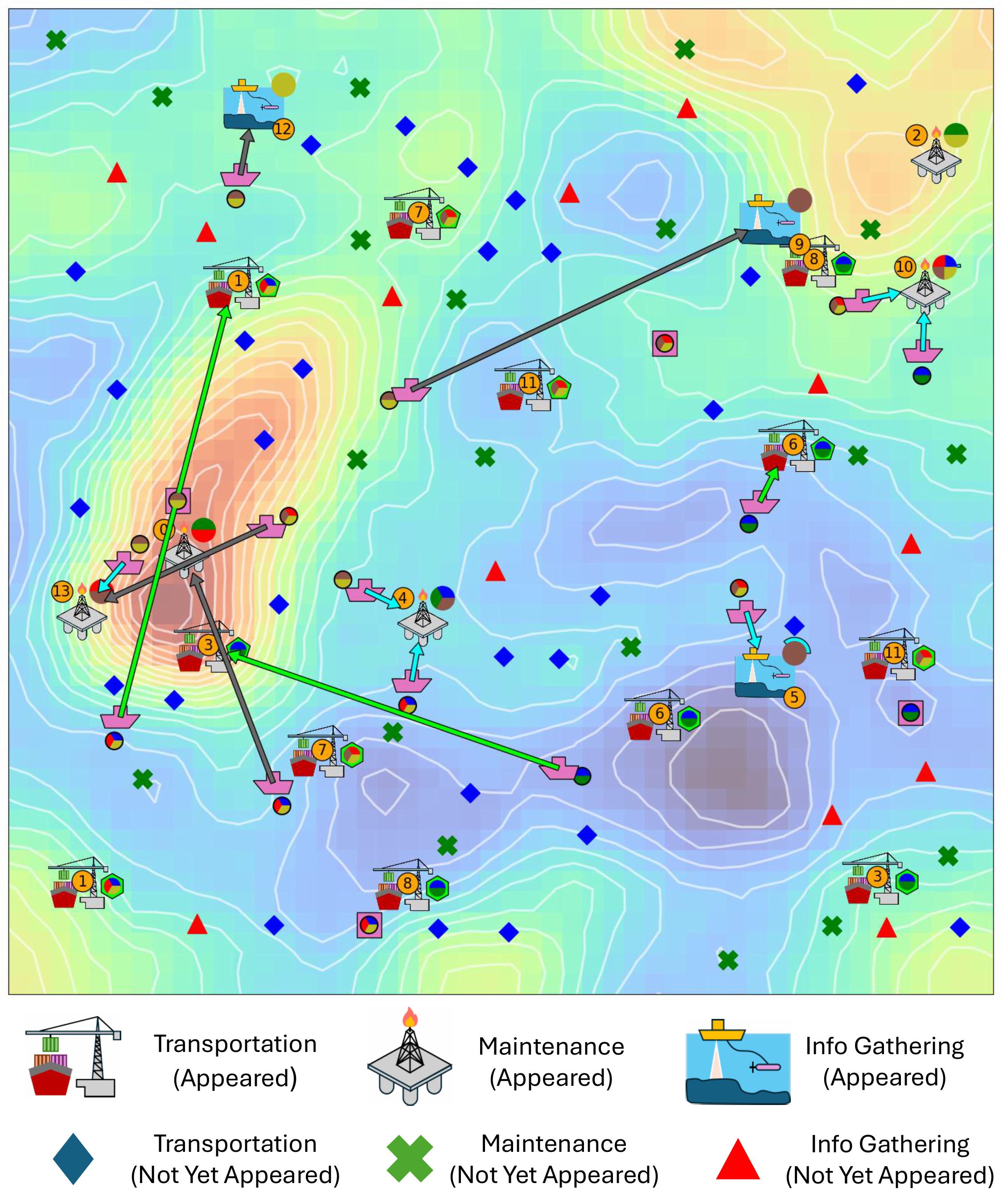}
    \caption{\textbf{Our task allocation problem.} The base station of each robot species is marked by a pink square. The capability vectors of robot species and the capability requirements of tasks are shown as color palettes in pink squares and next to robot and task icons, where each color represents a specific capability. Gray, lime and cyan arrows indicate the task location which the robot is en-route to executing, transporting items towards, and ready for (or under) execution, respectively. Background colors, from blue (low) to red (high), show local uncertainty levels that are unknown to robots.
    White contour lines indicate locations with the same uncertainties. 
    }
    \label{fig:demonstration}
    \vspace{-4mm}
\end{figure} 
Autonomous Surface Vehicles (ASVs) in maritime environments are typically subject to the influences of spatially and temporally varying weather and sea conditions \cite{knauss2016introduction}.   
When operating in areas covered by 
storms, the motions of ASVs can be substantially influenced by storm-induced wind, waves and ocean currents \cite{fossen2011handbook}, such that it can take a longer time to travel to designated locations and perform assigned tasks.
In addition,  communication quality throughout the environment may degrade significantly due to rain-induced signal attenuation and antenna misalignment from significant motion disturbances in stormy regions.
In this work, we focus on the task allocation problem of ASVs under the effects of spatio-temporal maritime environmental disturbances like storms.

Given that autonomous vehicles can prevent the human-induced errors and risks that exist in human-controlled vehicles, ASVs have been increasingly deployed for various kinds of maritime missions \cite{vagale2021path}.
To evaluate the performance of task allocation solutions over different kinds of tasks, we consider important maritime missions with heterogeneous task requirements. 
As shown in Fig. \ref{fig:demonstration}, our problem includes information gathering tasks that are stationary single-robot tasks \cite{liu2019intelligent,wei2020particle,mahato2023consensus}, maintenance tasks that require multi-robot collaboration \cite{fu2022robust,dai2025heterogeneous}, and transportation tasks that are similar to those in pickup-and-delivery scenarios \cite{liu2019task,chen2021integrated,agrawal2023rtaw}.
As our task allocation environment is vast and 
unstructured,
the paths of transportation tasks are simplified as straight lines connecting start and goal locations. 
Furthermore, our tasks have different appearance times, such that the task allocation policy needs to perform efficient re-planning when new tasks show up in the environment \cite{chen2021integrated,agrawal2023rtaw}.

Due to the communication issues induced by environmental uncertainties, task allocation methods that use centralized planners may experience robot action failures when robots don't receive commands from the planner. 
For decentralized methods, each robot makes individual decisions and can always act when available, but it can be challenging to perform multi-robot collaboration in a decentralized manner.     
We aim to take advantage of the action robustness of a decentralized method (due to individual decision making), and to foster cooperative behaviors among robots when necessary. Accordingly, we propose a decentralized decision making policy based on Deep Reinforcement Learning (DRL), which can adapt to heterogeneous tasks via trial-and-error learning \cite{sutton1998reinforcement}.
Since the observations of task and robot states support the estimation of travel and task execution times, which may deviate from the actual values under the influence of environmental disturbances, we design a policy model structure that can learn to be robust to erroneous estimation of task execution time. 
The contributions of this work are summarized as follows:

\begin{itemize}
    \item We formulate a multi-robot task allocation problem under the challenges of 
    unknown stochastic spatio-temporal uncertainties in maritime environments, and tasks of heterogeneous types with different appearance times.
    \item We design a decentralized policy network that handles deviated observations due to environmental uncertainties via an adaptive weighting module, and performs attention-based fusions for team-and-task-aware individual decision making. An episodic reward considering episode time length and statuses of heterogeneous tasks is used to train a Reinforcement Learning (RL) based policy with the proposed model.  
    \item The proposed RL based policy has been comprehensively evaluated against four baselines in scenarios with varying task type proportions, numbers of tasks, robot species and robots per species. Experiment results show that the proposed method demonstrates superior performance over all baselines.
    \item Our approach is freely available at \url{https://github.com/RobustFieldAutonomyLab/Spatio_Temporal_MRTA}.
\end{itemize}

\section{Related Works}
\label{sec:related works}   

Many existing works on multi-robot collaboration model uncertainties with basic probability distributions.
Zhou et al. \cite{zhou2018approximation} proposed a sequential greedy algorithm for a multi-vehicle mobility-on-demand problem with path-related arrival time uncertainty.    
Peltzer et al. \cite{peltzer2020stt} studied a multi-agent path finding problem in warehouse scenarios where agent travel time stochasticity follows a gamma distribution.
Choudhury et al. \cite{choudhury2022dynamic} proposed a stochastic conflict-based task allocation method for multi-drone package delivery missions with travel time uncertainty based on Epanechnikov and normal distributions.  
Park et al. \cite{park2018robust} targeted a multi-robot collaboration problem under random robot failures modeled with the binomial distribution and proposed a distributed stochastic optimal control solution.  
Huo et al. \cite{huo2024self} introduced agent observation uncertainties, modeled with Dirichlet distributions, into a multi-robot coalition formation problem and addressed it with a belief updating method.
Our work presents a novel study on a multi-robot task allocation problem under spatio-temporal environmental uncertainties that affect robot travel and task execution speed, communication quality, and observations of robot and task states. 

Our problem in general falls into the ST-MR-TA category (single-task robots, have multi-robot tasks, and may need time-extended assignments) \cite{gerkey2004formal}. Specifically, it contains heterogeneous types of tasks across different categories of task allocation problems, which involve stationary single-robot tasks, multi-robot collaboration tasks, and pickup-and-delivery style tasks with simplified paths. Relevant prior works addressing each of these three classes of tasks are summarized in the three respective paragraphs that follow.

(1) \textit{Stationary Single-Robot Tasks}:
Ma et al. \cite{ma2021unsupervised} proposed a multi-ASV task allocation policy based on self-organizing maps for single-robot maritime tasks in constrained environments.
Wei et al. \cite{wei2020particle} formulated the multi-robot allocation problem as multiple Traveling Salesman Problems (TSPs), applying a multi-objective particle swarm optimization solution.
Banfi et al. \cite{banfi2022hierarchical} formulated a multi-robot heterogeneous area inspection mission as a vehicle routing problem, and developed a hierarchical planner which adopts a Graph Neural Network for allocation cost estimations.
Mahato et al. \cite{mahato2023consensus} proposed a market-based method which demonstrated energy-efficient performance on a distributed multi-robot task allocation problem.
Samiei et al. \cite{samiei2023distributed} converted a multi-task assignment problem for multi-agent systems into a linear sum assignment problem for local planning, and proposed a solution based on the Hungarian algorithm.

(2) \textit{Multi-Robot Collaboration Tasks}: Fu et al. \cite{fu2022robust} developed a Mixed-Integer Programming framework for solving task assignments and scheduling 
for multi-robot tasks with uncertainties on robot team capabilities.   
Dai et al. \cite{dai2025heterogeneous} proposed a Reinforcement Learning based policy that achieves comparable makespan to optimization based methods with efficient computation time and good scalability towards large-scale problems.  
Babincsak et al. \cite{babincsak2023ant} formulate a multi-objective optimization problem for multi-robot collaboration that considers the costs of all robots' paths and the maximum individual cost, which can be solved with Ant Colony Optimization based methods.
Neville et al. \cite{neville2021interleaved} introduced a heuristic based graph search approach to addressing requirements of multi-robot task scheduling and motion planning, and evaluates heuristics on task allocation quality as well as plan execution time.
Huang et al. \cite{huang2025novel} targeted the multi-robot collaborative scheduling problem in manufacturing scenarios and utilized genetic algorithm and constraint programming to deal with optimizations at different levels.
Optimization-based methods like \cite{fu2022robust} have exponential complexity, and it can take hundreds of seconds to generate solutions for large-scale problems \cite{dai2025heterogeneous}.

(3) \textit{Pickup-and-Delivery Tasks}: Ma et al. \cite{ma2017lifelong} introduced the multi-agent pickup and delivery (MAPD) problem as an extension of a multi-agent path finding problem with continual new incoming tasks, and performed online task allocation with a Hungarian-based method.
Liu et al. \cite{liu2019task} formulated the MAPD problem as a Traveling Salesman Problem, computing task sequences for each agent with a TSP solver \cite{helsgaun2017extension}.
Chen et al. \cite{chen2021integrated} aimed at the MAPD problem in warehouses and sortation centers, and presented a regret-based solution that iteratively assigns tasks to robots with minimum marginal costs.   
Agrawal et al. \cite{agrawal2023rtaw} proposed a Reinforcement Learning based method that trains a policy to minimize the total travel delay during task execution and showed improvement over the regret-based approach.  
Makino et al. \cite{makino2024online} introduced task deadlines into the multi-agent pickup and delivery problem, and proposed algorithms that consider both task costs and deadlines, and strategies for swapping 
tasks among robots.

\section{Problem Formulation}
\label{sec:problem formulation}
\subsection{Reinforcement Learning}
The task allocation problem can be formulated as a Markov Decision Process $(\mathcal{S},\mathcal{A},P,R,\gamma)$ \cite{dai2025heterogeneous,agrawal2023rtaw}, where $\mathcal{S}$ and $\mathcal{A}$ are the sets of environment states and agent actions, which are determined by the set of active tasks.
$P(s'|s,a)$ is the state transition function reflecting the environment dynamics. $R(s,a)$ is the reward function, and $\gamma\in [0,1]$ is the discount factor.
At each time step $t$, given the observation of the current state $s_t$, each agent chooses an action $a_t$, which causes the transition to the state $s_{t+1}\sim P(\cdot|s_t,a_t)$ and receiving the reward $r_{t+1}=R(s_{t+1},a_{t+1})$.  
The value function $V^\pi(s)$ is defined as the expected cumulative reward of taking actions according to the policy $\pi$ when starting from state $s$.
\begin{equation}
  \label{eq:expected return}
  V^\pi(s) = \mathbb{E}_\pi[\sum_{k=0}^\infty \gamma^k r_{t+k+1}| s_t=s]
\end{equation}
An optimal policy $\pi_*$ maximizes $V^\pi(s)$, and the resulting optimal value function $V^{\pi_*}(s)$ satisfies the Bellman optimality equation \eqref{eqn:Bellman optimality equation}.
\begin{equation}
    \label{eqn:Bellman optimality equation}
        V^{\pi_*}(s) = \max_a\mathbb{E}[r_{t+1}+\gamma V^{\pi_*}(s_{t+1})|s_t=s,a_t=a]
\end{equation}

\subsection{Heterogeneous Robots and Tasks}
Similar to relevant prior works \cite{dai2025heterogeneous,prorok2017impact,fu2022robust}, we assume that the whole robot team can be divided into multiple robot species, and robots within a certain species share the same capability vector.
The location of a base station indicates the initial position of robots of the corresponding species, and all robots need to return to their base stations to complete the episode.
As shown in Fig. \ref{fig:demonstration}, in this work we consider three heterogeneous types of tasks: 

(1) \textit{Transportation}: Its requirement is the same as the capability vector of one specific robot species and can only be performed by one of those robots. The executing robot needs to travel from the transportation start location to goal location to complete the task.

(2) \textit{Maintenance}: It requires that all capabilities in its requirement need to be satisfied by the cumulative capabilities of the group of executing robots, and the execution process doesn't start until all robots reach the task location.

(3) \textit{Information Gathering}: Its requirement contains only one capability and can be performed by any robot that owns the capability upon reaching the task location. 

\subsection{Stochastic Spatio-Temporal Environmental Uncertainty}
To study the policy performance under the high variability of maritime environments, we construct a computationally efficient stochastic spatio-temporal uncertainty simulation inspired by fundamental atmospheric transport principles. 

We use Gaussian ellipses to model localized high uncertainty regions like storm cores, which are commonly used as simplified representations of mesoscale atmospheric cells \cite{wallace2006atmospheric}.
The uncertainties are transported by a slowly varying wind vector, approximating the classical advection equation widely used to model atmospheric scalar transport \cite{vallis2017atmospheric}.
A multi-octave Gaussian-smoothed noise process is used to generate large-scale correlated background variability, approximating the multi-scale spatial structure commonly observed in atmospheric phenomena \cite{frisch1995turbulence}.
After superimposing the components mentioned above, we normalize the uncertainty values to vary between 0 and 1.

We assume that the amounts of delay in local travel speed and task execution speed of each robot are proportional to its local uncertainty level.
In addition, the probabilities of communication failures among robots as well as with centralized planners (used by some of the baseline methods in experiments) are proportional to their local uncertainty levels.
We also assume that centralized planners and robot onboard decentralized planners are unaware of the uncertainty value distributions over the whole working environment.

\section{Methodology}
\label{sec:methodology} 

\subsection{Observation}
When a robot is at a decision-making moment, it is given the state observation of itself, other robots the information of which are successfully obtained at the current moment, as well as active tasks. 
\textbf{\textit{C}} is the robot capability vector; 
\textbf{\textit{P}} is the relative position of a task (for a transportation task it is the start position) or another robot with respect to the ego robot;
$\textbf{\textit{P}}_{goal}$ is the relative goal position of a transportation task (for other tasks it is the same as \textbf{\textit{P}});
\textbf{\textit{R}} is the task requirement vector;
\textbf{\textit{S}} reflects the capabilities in \textbf{\textit{R}} that have not yet been satisfied by assigned robots;
\textit{I} indicates whether \textbf{\textit{R}} is completely satisfied;
$t_{wait\rightarrow T}$ is the difference between current time and the time when the robot reached the task location;
$\hat{t}_{travel\rightarrow T}$ and $\hat{t}_{travel\rightarrow T}$ represent the estimated travel time to assigned task $T$ when there is no interference from the environmental uncertainty. 
Before passing into the encoder, robot observations are concatenated with the summary information of assigned tasks.

\begin{figure*}
    \centering
    \includegraphics[width=0.95\linewidth]{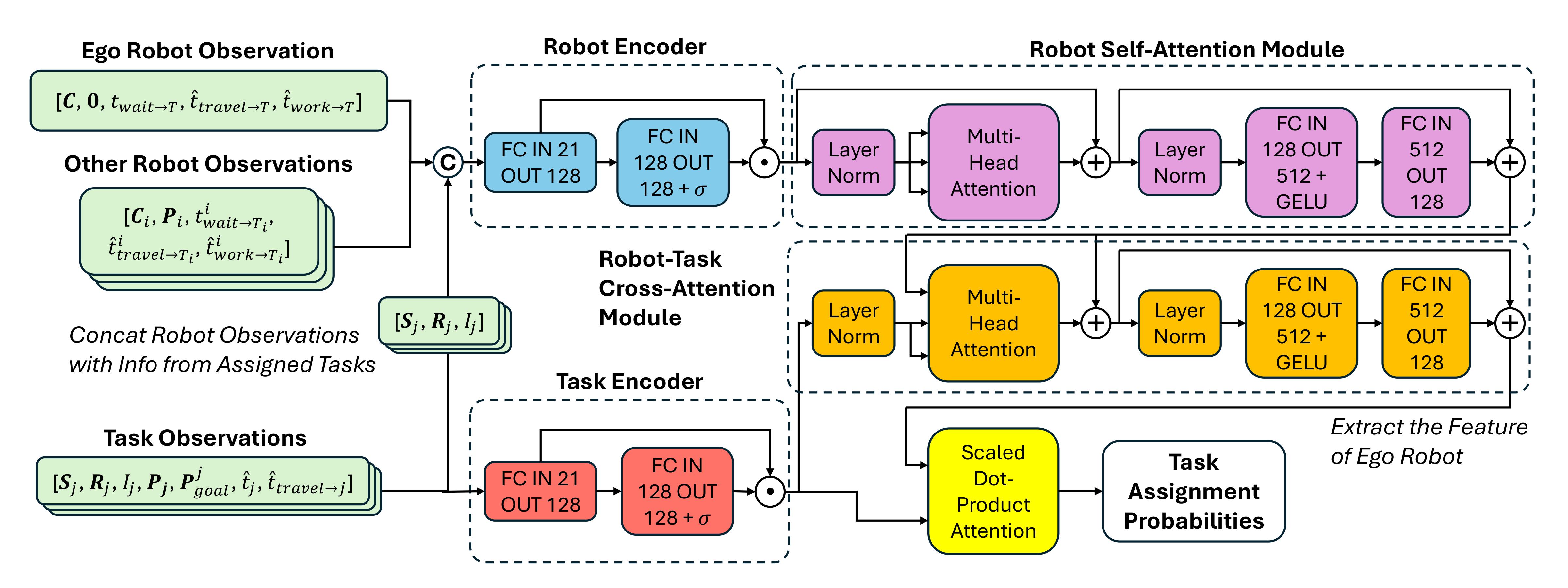}
    \caption{\textbf{Policy network architecture.} Observations of robots and tasks are passed through corresponding gated encoders to obtain robust embedded features. Then a self-attention module and a cross-attention module are used to fuse the ego robot feature with that of other robots and tasks to obtain a comprehensive understanding. The policy outputs a probability vector that reflects the attentions of the ego robot on all available tasks.}
    \label{fig:networks}
    \vspace{-6mm}
\end{figure*}

\subsection{Reward}
The objective in this problem is to obtain a policy that minimizes the total makespan when each agent performs individual decision making, and it is important to reduce myopic and greedy actions that only consider the current situation.
Hence we assume that $\gamma=1.0$ and $r=0$ until the end of an episode is reached, where an episodic reward $r=r_{\text{episode}}$ as shown below is given, such that the value function at all intermediate states in an episode would be equal to $r_{\text{episode}}$ and thus they contribute equally to the final outcome.
\begin{equation}
    \begin{aligned}
    r_{\text{episode}} &= -t_{\text{episode}} -\sum_{T_j\in\mathbb{F}}(I(T_j\notin\mathbb{T}) \cdot ||\Sigma_{i\in A(T_j)}C_i||/||R_j|| \\
    &+I(T_j\in\mathbb{T}))\cdot r_{\alpha} - |T_l\notin\mathbb{F}|\cdot r_{\beta}
    \end{aligned}
    \label{eq:reward}
\end{equation}

$t_{\text{episode}}$ is the total time step of the episode, which is terminated if all tasks have been completed and all robots have returned to base stations, or a given max time step is reached; 
$I(\cdot)$ is the indicator function; 
$\mathbb{T}$ is the set of transportation tasks; 
$A(\cdot)$ is the set of assigned robots to a task; 
$F$ is the set of finished tasks at the end of an episode.
$r_\alpha$ and $r_\beta$ are the penalties for finished and unfinished tasks. 

Under the episodic reward setting, similar to the approach in \cite{dai2025heterogeneous}, we adopt a Monte-Carlo policy gradient method for policy training, which aims at maximizing the objective function $R(\tau)$ that is the expected reward of an episode $\tau$.
\begin{equation}
    J(\theta)=\mathbb{E}_{\tau\sim\pi_{\theta}}[R(\tau)],\ \tau=\{(i_k,o_k,a_k)|k=1,\dots,K\}
\end{equation}
In our problem setting, $R(\tau)$ is the episodic reward shown in Equation \eqref{eq:reward}, and an episode is a set of decision making experiences where $i$, $o$ and $a$ in each experience represent the index of the decision making agent, its observations and action.
The gradient for policy updates, $\nabla_{\theta}J(\theta)$, is shown as follows.
A baseline value $b$ is used to reduce variance, which is computed as the average episodic reward of multiple runs of a training episode.
\begin{equation}
\label{eq:policy gradient}
    \nabla_\theta J(\theta)=\mathbb{E}[\sum_{k=1}^K\nabla_{\theta}\log\pi_{\theta}(a_k|o_k)(R(\tau)-b)]
\end{equation}

\subsection{Policy Model Structure}
Given that observations contain estimated future travel and task execution information, which may deviate from the actual values that are influenced by environmental uncertainties, we adopt robot and task encoders with gate units to mitigate the effect of deviated estimations on embedded features, which is shown in Equation \eqref{eq:encoder}.
\begin{equation}
\label{eq:encoder}
    \mathcal{F}=\ \mathcal{F_{\text{raw}}}\odot\sigma(W_2\mathcal{F_{\text{raw}}}+b_2),\ \mathcal{F_{\text{raw}}}=W_1\mathcal{O}+b_1
\end{equation}
Raw features $\mathcal{F_{\text{raw}}}$ are obtained by performing linear projections on observations $\mathcal{O}$, then we compute adaptive weights of $\mathcal{F_{\text{raw}}}$ via another linear unit and sigmoid function $\sigma(\cdot)$, which are applied to $\mathcal{F_{\text{raw}}}$ via element-wise multiplication $\odot$ to produce robust embeddings $\mathcal{F}$. 

During the decentralized decision making, each robot needs to independently consider the status of other robots and active tasks to balance the demands such as efficiently reducing the remaining task work amount, avoiding wasting time on tasks that other robots may do better, and helping with cooperative tasks where others have been waiting.
Hence, we perform feature fusions in the robot self-attention module and robot-task cross-attention module, which are based on the scaled dot-product attention function $\text{Attention}(\cdot)$ and the multi-head attention function $\text{MHA}(\cdot)$, which are defined in \cite{vaswani2017attention} and are shown below.
\begin{equation}
\label{eq:attention}
\text{Attention}(Q,K,V)=\text{Softmax}(\frac{QK^T}{\sqrt{d_k}})V
\end{equation}
\begin{equation}
\label{eq:MHA}
    \begin{array}{c}
    \text{MHA}(Q,K,V)=C(H_1,\dots,H_h)W^O, \\[5pt]
    \text{where }H_i=\text{Attention}(QW^Q_i,KW^K_i,VW^K_i)
    \end{array}
\end{equation}

\begin{figure*}[th]
    \centering
    \begin{minipage}{0.32\linewidth}
    \includegraphics[width=\linewidth]{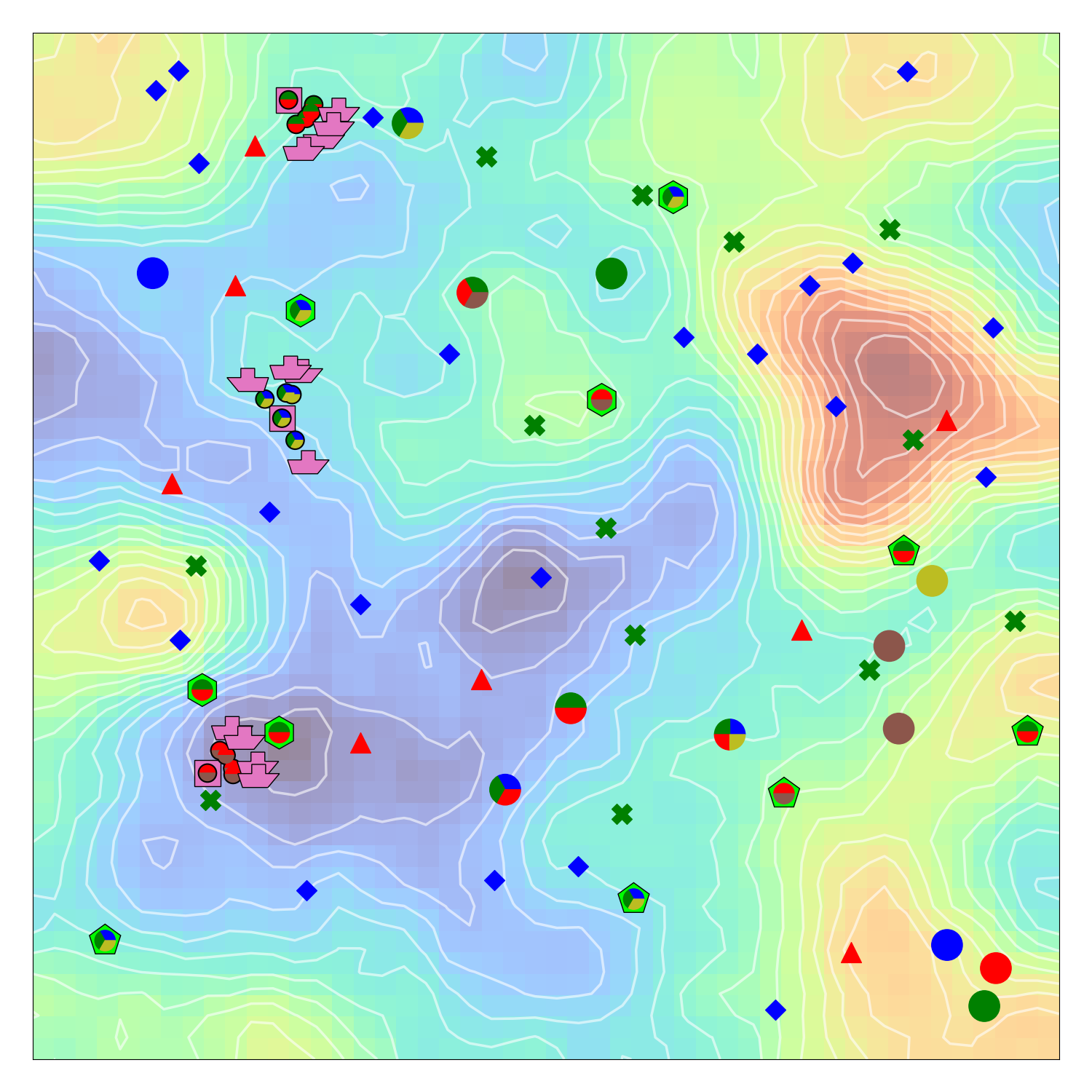}
    \subcaption{$m=50$, $n=3$, $k=4$}
    \end{minipage}
    \begin{minipage}{0.32\linewidth}
    \includegraphics[width=\linewidth]{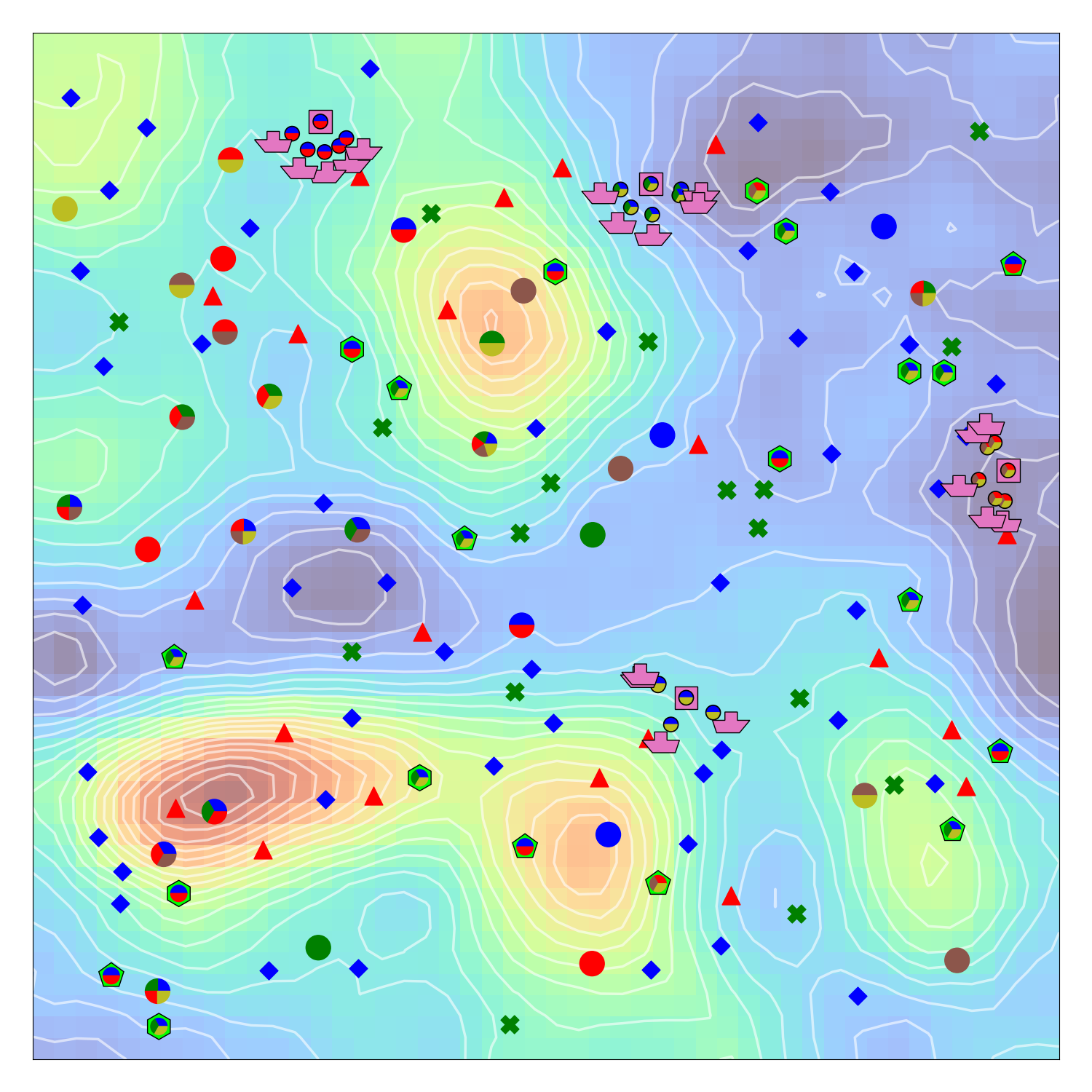}
    \subcaption{$m=100$, $n=4$, $k=5$}
    \end{minipage}
    \begin{minipage}{0.32\linewidth}
    \includegraphics[width=\linewidth]{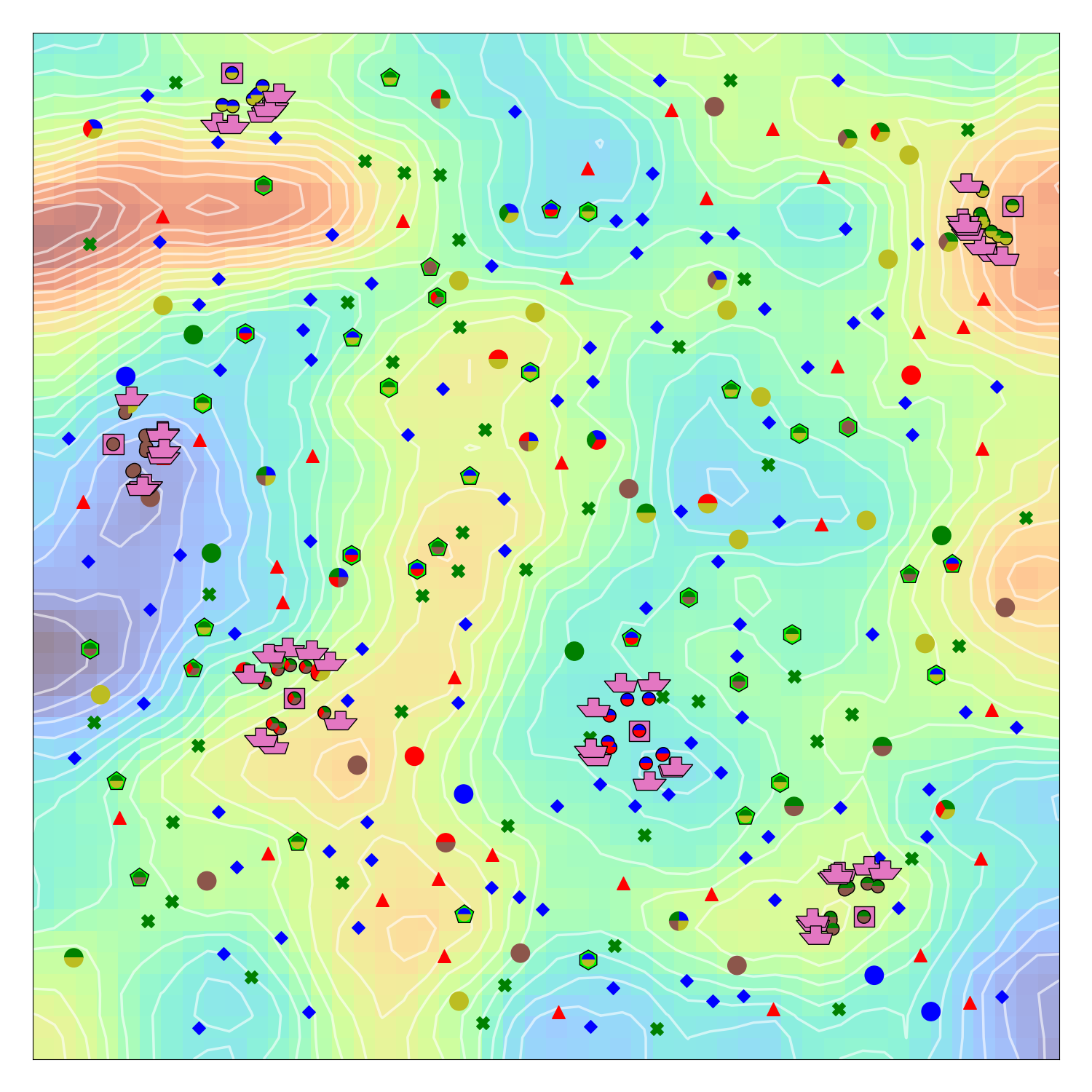}
    \subcaption{$m=200$, $n=6$, $k=8$}
    \end{minipage}
    \vspace{2mm}
    \caption{\textbf{Examples of initial scenes in experiment episodes.} The capability requirements of active tasks at the current moment are shown as color palettes, while hexagons and pentagons with lime colors indicate the start and goal locations of transportation tasks. Markers for tasks that have not yet appeared are the same as those in Fig. \ref{fig:demonstration}.}
    \vspace{-5mm}
    \label{fig:example episodes}
\end{figure*}

In the robot self-attention module, $Q$, $K$ and $V$ are the same and come from the embedded robot features such that each robot feature is fused with that of other robots according to the learned attentions at the current time step, and the module outputs team-status-aware robot features.
In the robot-task cross-attention module, $Q$ are the team-status-aware robot features, $K$ and $V$ come from the embedded task features, then the learned attentions to each task serve as the weights to fuse task features into the team-status-aware robot features, resulting in comprehensively fused features. 
The policy outputs task assignment probabilities via a scaled dot-product attention function given the final fused feature of the ego robot and task features.
A categorical distribution on the task assignment probabilities is used to determine the next task during the training process, while in experiments the task with the maximum probability is chosen.

\subsection{Model Training}
\label{sec:training}
\begin{figure}[h]
    \centering
    \includegraphics[width=\linewidth]{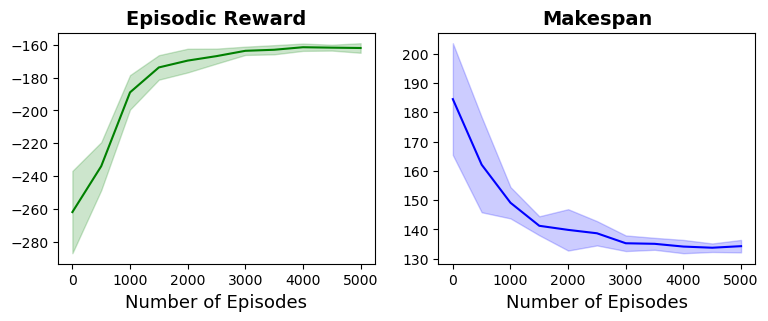}
    \caption{\textbf{Training performance.} The solid line and bandwidth represent the mean and standard deviation of evaluation results over ten models from ten different training trials.}
    \label{fig:training curves}
    \vspace{-5mm}
\end{figure}

We performed the training process on an Nvidia RTX 3090 GPU.
The list of training hyperparameters we use is shown in Table \ref{tab:hyperparameters}, where ep, $m$, $n$, $k$ stand for episode, the number of tasks, robot species, and robots per species.
To enable the trained policy to generalize to a variety of scenarios with different map sizes where numbers and densities of tasks and robots vary, the location coordinates in simulation environments are normalized between 0 and 1.
Each episode is initialized with random $m$, $n$ and $k$ within the range, random task and robot base station locations, as well as random batches of appearance times for tasks.
We evaluate the learning model after every 500 training episodes, and the training performance is summarized in Fig. \ref{fig:training curves}, which shows that the model performance converges in 5000 episodes.

\begin{table}[h!]
    \centering
    \caption{\textbf{Training hyperparameters}.
    }
    \label{tab:hyperparameters}
    \small
    \begin{tabular}{>{\centering\arraybackslash}m{2.0cm}>{\centering\arraybackslash}m{1.2cm}|>{\centering\arraybackslash}m{2.0cm}>{\centering\arraybackslash}m{1.2cm}}
        \toprule
        range of $m$ &$[75,150]$  & learning rate & $1e-5$ \\ [1ex]
        range of $n$ & $[3,5]$ & batch size & $1024$ \\ [1ex]
        range of $k$ & $[4,6]$ & eval ep num & $256$ \\ [1ex]
        max ep time & 200 & total ep num & 5000 \\
        \bottomrule
    \end{tabular}
\end{table}

\section{Experiments}
\label{sec:experiments}

\subsection{Experiment Settings}

To comprehensively evaluate the performance of our proposed policy in this problem, we consider four additional methods which serve as comparative baselines and are introduced in Sec. \ref{sec:baselines}, and perform multiple sets of simulation experiments with varying task type proportions, $m$, $n$ and $k$, where each set of experiments (each column in Table \ref{tab:exp results}) contains 100 experiment episodes.
Each experiment episode is generated in the same manner as described in Sec. \ref{sec:training}, and the visualizations of example initial environment configurations are shown in Fig. \ref{fig:example episodes}.
Note that the initial positions of all robots of the same species are the same as the location of the corresponding base station, and we show robot icons surrounding base stations in Fig. \ref{fig:example episodes} just to indicate the number of robots per species.

\subsection{Baselines}
\label{sec:baselines}
Given that tasks may appear at different moments in our problem, we choose the following baseline methods which are able to perform efficient online re-planning, and make necessary adaptations to our problem with heterogeneous types of tasks.
The cost function used by the baselines for evaluating each robot-to-task assignment is computed as the sum of estimated travel time and estimated task time. 

(1) \textit{Heuristic Based Task Allocation  (HBTA)}: Greedy heuristic methods have been widely implemented as evaluation baselines \cite{dai2025heterogeneous,agrawal2023rtaw,banfi2022hierarchical}. To adapt to multi-robot tasks in our problem, we utilize the idea of task decomposition \cite{korsah2013comprehensive} and divide each cooperative task into single capability requirement slots.
Then, for every capability requirement slot in each task, a min cost heap is constructed for all available robots, and the robot with minimum cost and without any assignment conflicts is selected. 

(2) \textit{Decentralized HBTA (D-HBTA)}: In contrast to HBTA, for which a centralized planner considers the costs of all available robots, we also include a decentralized version that assigns a task to the ego robot only based on its marginal costs with respect to all active tasks.

(3) \textit{Consensus-Based Bundle Algorithm (CBBA)}: Based on the method in \cite{choi2009consensus}, a distributed combinatorial auction process is formulated, where each robot iteratively constructs a task bundle by greedily selecting assignments that minimize its marginal cost under feasibility constraints, followed by a consensus phase where neighboring robots exchange bid information to resolve conflicts and achieve agreement on task ownership.
Similar to the idea in \cite{braquet2021greedy}, multiple robots are allowed to win the same cooperative task until all capability requirements of the task are satisfied.

(4) \textit{Hungarian Algorithm (Hungarian):} In multi-robot task allocation problems, as illustrated in \cite{msala2023new}, given a bipartite graph between robots and tasks with an associated cost matrix where each entry represents the cost of a specific robot-to-task assignment, the Hungarian algorithm computes a one-to-one assignment that minimizes the total cost by solving the linear sum assignment problem.
For cooperative tasks, we also adopt the task decomposition technique mentioned above such that the one-to-one assignment constraint in the Hungarian Algorithm is satisfied.

\subsection{Results}
\begin{table}[ht!]
    \centering
    \caption{\textbf{Makespan data in experiments}. Each entry shows the mean and standard deviation of results from 100 experiments.
    }
    \label{tab:exp results}
    \vspace{1mm}
    \subcaption{$50\%$ info gathering, $25\%$ maintenance, $25\%$ transportation}
    \begin{tabular}{>{\centering\arraybackslash}m{1cm}>{\centering\arraybackslash}m{0.78cm}>{\centering\arraybackslash}m{0.78cm}>{\centering\arraybackslash}m{0.78cm}>{\centering\arraybackslash}m{0.78cm}>{\centering\arraybackslash}m{0.78cm}>{\centering\arraybackslash}m{0.78cm}}
         \toprule
          $m$ & 50 & 100 & 200 & 200 & 400 & 400  \\ [0.5ex] $n$ & 3 & 4 & 4 & 6 & 7 & 10 \\ [0.5ex] $k$ & 4 & 5 & 8 & 8 & 10 & 10 \\ \midrule
        \multirow{2}{*}{RL (Ours)} & {\bf 75.92} & {\bf 94.72} & {\bf 109.50} & {\bf 82.10} & {\bf 109.26} & {\bf 87.25} \\ [0.2ex]
        & {\scriptsize $\bf \pm 22.45$} & {\scriptsize $\bf \pm 30.29$} & {\scriptsize $\bf \pm 28.40$} & {\scriptsize $\bf \pm 18.59$} & {\scriptsize $\bf \pm 32.39$} & {\scriptsize $\bf \pm 25.05$} \\ [1ex]
        \multirow{2}{*}{HBTA} & 101.78 & 119.54 & 129.20 & 96.38 & 121.33 & 91.44 \\ [0.2ex]
        & {\scriptsize $\pm 40.44$} & {\scriptsize $\pm 39.43$} & {\scriptsize $\pm 31.49$} & {\scriptsize $\pm 26.12$} & {\scriptsize $\pm 36.83$} & {\scriptsize $\pm 24.72$} \\ [1ex]
        \multirow{2}{*}{D-HBTA} & 85.79 & 108.95 & 129.36 & 94.91 & 122.66 & 93.57 \\ [0.2ex]
        & {\scriptsize $\pm 28.88$} & {\scriptsize $\pm 30.30$} & {\scriptsize $\pm 33.68$} & {\scriptsize $\pm 25.84$} & {\scriptsize $\pm 38.05$} & {\scriptsize $\pm 23.42$} \\ [1ex]
        \multirow{2}{*}{CBBA} & 90.53 & 116.74 & 139.68 & 105.75 & 134.94 & 105.25 \\ [0.2ex]
        & {\scriptsize $\pm 27.43$} & {\scriptsize $\pm 29.98$} & {\scriptsize $\pm 35.41$} & {\scriptsize $\pm 27.12$} & {\scriptsize $\pm 30.78$} & {\scriptsize $\pm 30.36$}
        \\ [1ex]
        \multirow{2}{*}{Hungarian} & 108.95 & 134.47 & 153.75 & 109.20 & 136.54 & 99.49 \\ [0.2ex]
        & {\scriptsize $\pm 38.12$} & {\scriptsize $\pm 38.99$} & {\scriptsize $\pm 41.47$} & {\scriptsize $\pm 30.26$} & {\scriptsize $\pm 38.36$} & {\scriptsize $\pm 27.00$} \\
        \bottomrule
    \end{tabular}
    \vspace{3mm}
    \subcaption{$25\%$ info gathering, $50\%$ maintenance, $25\%$ transportation}
    \begin{tabular}{>{\centering\arraybackslash}m{1cm}>{\centering\arraybackslash}m{0.78cm}>{\centering\arraybackslash}m{0.78cm}>{\centering\arraybackslash}m{0.78cm}>{\centering\arraybackslash}m{0.78cm}>{\centering\arraybackslash}m{0.78cm}>{\centering\arraybackslash}m{0.78cm}}
         \toprule
          $m$ & 50 & 100 & 200 & 200 & 400 & 400  \\ [0.5ex] $n$ & 3 & 4 & 4 & 6 & 7 & 10 \\ [0.5ex] $k$ & 4 & 5 & 8 & 8 & 10 & 10 \\ \midrule
        \multirow{2}{*}{RL (Ours)} & {\bf  87.76} & {\bf 113.85} & {\bf 134.21} & {\bf 96.12} & {\bf 129.31} & {\bf 98.31} \\ [0.2ex]
        & {\scriptsize $\bf \pm 26.59$} & {\scriptsize $\bf \pm 32.47$} & {\scriptsize $\bf \pm 34.42$} & {\scriptsize $\bf \pm 21.87$} & {\scriptsize $\bf \pm 32.79$} & {\scriptsize $\bf \pm 25.75$} \\ [1ex]
        \multirow{2}{*}{HBTA} & 111.90 & 134.95 & 152.48 & 114.97 & 136.70 & 104.02 \\ [0.2ex]
        & {\scriptsize $\pm 39.96$} & {\scriptsize $\pm 38.27$} & {\scriptsize $\pm 37.58$} & {\scriptsize $\pm 33.19$} & {\scriptsize $\pm 36.19$} & {\scriptsize $\pm 32.27$} \\ [1ex]
        \multirow{2}{*}{D-HBTA} & 100.95 & 136.36 & 158.85 & 117.20 & 142.27 & 109.67 \\ [0.2ex]
        & {\scriptsize $\pm 32.19$} & {\scriptsize $\pm 39.39$} & {\scriptsize $\pm 41.43$} & {\scriptsize $\pm 31.62$} & {\scriptsize $\pm 34.82$} & {\scriptsize $\pm 31.86$} 
        \\ [1ex]
        \multirow{2}{*}{CBBA} & 103.44 & 136.37 & 160.09 & 125.39 & 150.66 & 116.54 \\ [0.2ex]
        & {\scriptsize $\pm 26.60$} & {\scriptsize $\pm 33.12$} & {\scriptsize $\pm 38.33$} & {\scriptsize $\pm 35.92$} & {\scriptsize $\pm 32.03$} & {\scriptsize $\pm 32.47$} 
        \\ [1ex]
        \multirow{2}{*}{Hungarian} & 123.61 & 154.31 & 166.75 & 133.86 & 152.19 & 123.74 \\ [0.2ex]
        & {\scriptsize $\pm 41.59$} & {\scriptsize $\pm 37.64$} & {\scriptsize $\pm 32.87$} & {\scriptsize $\pm 35.90$} & {\scriptsize $\pm 28.19$} & {\scriptsize $\pm 36.50$}
        \\
        \bottomrule
    \end{tabular}
    \vspace{3mm}
    \subcaption{$25\%$ info gathering, $25\%$ maintenance, $50\%$ transportation}
    \begin{tabular}{>{\centering\arraybackslash}m{1cm}>{\centering\arraybackslash}m{0.78cm}>{\centering\arraybackslash}m{0.78cm}>{\centering\arraybackslash}m{0.78cm}>{\centering\arraybackslash}m{0.78cm}>{\centering\arraybackslash}m{0.78cm}>{\centering\arraybackslash}m{0.78cm}}
         \toprule
          $m$ & 50 & 100 & 200 & 200 & 400 & 400  \\ [0.5ex] $n$ & 3 & 4 & 4 & 6 & 7 & 10 \\ [0.5ex] $k$ & 4 & 5 & 8 & 8 & 10 & 10 \\ \midrule
        \multirow{2}{*}{RL (Ours)} & {\bf 76.53} & {\bf 88.27} & {\bf 106.01} & {\bf 81.73} & {\bf 105.19} & {\bf 80.05} \\ [0.2ex]
        & {\scriptsize $\bf \pm 23.60$} & {\scriptsize $\bf \pm 20.06$} & {\scriptsize $\bf \pm 26.20$} & {\scriptsize $\bf \pm 24.33$} & {\scriptsize $\bf \pm 28.91$} & {\scriptsize $\bf \pm 17.01$} \\ [1ex]
        \multirow{2}{*}{HBTA} & 99.21 & 114.43 & 125.64 & 92.54 & 122.00 & 88.54 \\ [0.2ex]
        & {\scriptsize $\pm 35.98$} & {\scriptsize $\pm 35.64$} & {\scriptsize $\pm 34.09$} & {\scriptsize $\pm 24.61$} & {\scriptsize $\pm 29.55$} & {\scriptsize $\pm 23.22$} \\ [1ex]
        \multirow{2}{*}{D-HBTA} & 85.58 & 103.67 & 119.72 & 88.05 & 118.48 & 88.71 \\ [0.2ex]
        & {\scriptsize $\pm 28.90$} & {\scriptsize $\pm 23.84$} & {\scriptsize $\pm 31.32$} & {\scriptsize $\pm 23.21$} & {\scriptsize $\pm 28.62$} & {\scriptsize $\pm 22.51$} \\ [1ex]
        \multirow{2}{*}{CBBA} & 93.41 & 112.28 & 138.99 & 100.15 & 138.28 & 103.61 \\ [0.2ex]
        & {\scriptsize $\pm 28.54$} & {\scriptsize $\pm 23.28$} & {\scriptsize $\pm 42.37$} & {\scriptsize $\pm 24.25$} & {\scriptsize $\pm 32.89$} & {\scriptsize $\pm 28.92$} \\ [1ex]
        \multirow{2}{*}{Hungarian} & 112.65 & 125.54 & 145.95 & 100.91 & 137.69 & 97.32 \\ [0.2ex]
        & {\scriptsize $\pm 40.38$} & {\scriptsize $\pm 35.08$} & {\scriptsize $\pm 40.31$} & {\scriptsize $\pm 25.74$} & {\scriptsize $\pm 35.51$} & {\scriptsize $\pm 24.92$} \\
        \bottomrule
    \end{tabular}
    \vspace{-6mm}
\end{table}

The experiment results are summarized in Table \ref{tab:exp results}.
In general, our proposed policy shows scalability towards large robot teams (100 robots) and substantial numbers of tasks (400 tasks).
Compared to baseline methods, our method achieves the lowest makespan across different sets of experiments, indicating the robustness of our proposed policy in multi-robot task allocation scenarios under environmental uncertainties and heterogeneous task requirements.

Firstly, due to the influence of 
unknown environmental uncertainty, observation-based travel and task time estimation may deviate from the actual values to a varied extent, which compromises the accuracy of the assignment cost function used by baseline methods, leading to performance degradation. 
Our proposed policy is trained to deal with these deviations in observation via adaptive weighting of the raw features, such that the policy is less vulnerable to erroneous and/or corrupted observations.

In addition, communication failures due to the effect of environmental uncertainty impacts the evaluated methods to different degrees. 
For centralized methods, HBTA and Hungarian, a robot can perform an action only when it successfully sends an ego observation to and receives an action command from the centralized planner. Consequently, the number of acting robots is less than that of available robots. 
On the contrary, for decentralized methods, RL, D-HBTA and CBBA, all available robots can make action decisions on their own.
The number of acting robots can have a higher impact on algorithm performance when the total number of robots is small ($n=3$, $k=4$), and it can be seen that decentralized methods generally perform better than centralized ones.

Among decentralized methods, although CBBA enables every available robot to obtain an action, it may require multiple rounds of information exchange among robots, and thus the decision making process can be more vulnerable to communication failures. This results in higher makespan data compared to RL and D-HBTA, which require less communication among robots.
Since D-HBTA is a greedy approach that only considers the costs of the ego robot, our proposed policy that also takes the statuses of other robots into account leads to better team performance, especially in experiments with higher proportions of maintenance tasks that require multi-robot collaboration. 

\section{Conclusion}
\label{sec:conclusion}

In this work, we study a multi-robot task allocation problem in maritime environments with varying spatio-temporal uncertainties that are unknown to the robots, and heterogeneous tasks with different appearance times and requirements on robot capabilities.
We proposed a decentralized RL based policy with a model structure that can learn robust team-and-task-aware features for individual decision making. 
In the training process, we adopt an episodic reward that evaluates the episode time cost and heterogeneous task statuses to learn an efficient policy for various working scenarios.
The proposed policy shows superior performance in makespan when being evaluated against four baseline methods in simulation experiments that involve varying heterogeneous task type proportions, numbers of tasks, robot species and robots per species.
The proposed policy also demonstrates scalability to scenarios with 100 robots and 400 tasks. 
For future work, we may further study the policy performance on energy cost, and incorporate communication uncertainties with task sites into the problem.

\bibliographystyle{IEEEtran}
\bibliography{main}

\end{document}